\documentclass[conference]{IEEEtran}
\IEEEoverridecommandlockouts
\usepackage{times}
\usepackage{epsfig}
\usepackage{graphicx}
\usepackage{amsmath}
\usepackage{amssymb}
\usepackage{epsfig}
\usepackage{cite}
\usepackage{amsfonts}
\usepackage{algorithmic}
\usepackage{graphicx}
\usepackage{textcomp}
\usepackage[table]{xcolor}
\usepackage{lineno}
\usepackage{subfigure}
\usepackage{gensymb}
\usepackage{array,multirow}
\usepackage{amsmath}
\usepackage{amssymb}
\usepackage{makecell}
\usepackage{textcomp}
\usepackage{pifont}
\usepackage{graphics}
\def\BibTeX{{\rm B\kern-.05em{\sc i\kern-.025em b}\kern-.08em
    T\kern-.1667em\lower.7ex\hbox{E}\kern-.125emX}}
    \makeatletter
\newcommand{\linebreakand}{%
  \end{@IEEEauthorhalign}
  \hfill\mbox{}\par
  \mbox{}\hfill\begin{@IEEEauthorhalign}
}
\makeatother
\begin{document}

\title{Fusing Multiscale Texture and Residual Descriptors for Multilevel 2D Barcode Rebroadcasting Detection\\
\thanks{This work was partially funded by the MSCA-EU project PrintOut (Grant \#892757) and also by the National National Science Foundation of China (Grant \#62072313). Corresponding author: Changcheng Chen (cschen@szu.edu.cn)}
}

\author{\IEEEauthorblockN{Anselmo Ferreira}
\IEEEauthorblockA{\textit{Department of Information Engineering and Mathematics} \\
\textit{University of Siena}\\
Siena, Italy \\
anselmo.castelo@unisi.it}
\and
\IEEEauthorblockN{Changcheng Chen}
\IEEEauthorblockA{\textit{Shenzhen Key Laboratory of
Media Security} \\
\textit{Shenzen University}\\
Shenzhen, Guangdong, China \\
cschen@szu.edu.cn}
\linebreakand 

\IEEEauthorblockN{Mauro Barni}
\IEEEauthorblockA{\textit{Department of Information Engineering and Mathematics} \\
\textit{University of Siena}\\
Siena, Italy \\
barni@dii.unisi.it}
}

\maketitle

\begin{abstract}
Nowadays, 2D barcodes have been widely used for advertisement, mobile payment, and product authentication. However, in applications related to product authentication, an authentic 2D barcode can be illegally copied and attached to a counterfeited product in such a way to bypass the authentication scheme. In this paper, we employ a proprietary 2D barcode pattern and use multimedia forensics methods to analyse the scanning and printing artefacts resulting from the copy (rebroadcasting) attack. A diverse and complementary feature set is proposed to quantify the barcode texture distortions introduced during the illegal copying process. The proposed features are composed of global and local descriptors, which characterize the multi-scale texture appearance and the points of interest distribution, respectively. The proposed descriptors are compared against some existing texture descriptors and deep learning-based approaches under various scenarios, such as cross-datasets and cross-size. Experimental results highlight the practicality of the proposed method in real-world settings.
\end{abstract}


\section{Introduction}
The widespread availability of high quality printers and scanners has made it cheaper and easier for a counterfeiter to make illegal copies of existing barcodes and attach them to counterfeited products. As a matter of fact, product counterfeiting is becoming an unprecedented problem for the global trade, as, for example, the 2016 report from Frontier Economics \cite{icc2016} predicted that, in 2022, 991 billion dollars will be traded in fake products, with an estimate of 4.2 to 5.2 million jobs lost because of piracy.

In order to tackle such an issue, 2D barcode authentication techniques proposed in the literature have been divided into active and passive categories. Active techniques \cite{tkachenko2016two,ONO2016991} are based on specific printing materials that could not be replicated by forgers and are commonly very expensive. On the other hand, passive copy-proof techniques \cite{voloshynovskiy2016physical,8794824} exploit replication artifacts without using special imaging devices or barcode patterns, taking into account simple imaging devices such as smartphones.  Notwithstanding, there is still an open field for other cheap passive anti-counterfeiting techniques, especially with regard to new types of 2D barcodes. Another important aspect to be dealt in such application is the high cost (financial and procedural) of building big datasets in order to make deep learning suitable training data. Therefore, approaches that could learn well from a small amount of training data are crucial to build efficient and cheap authentication approaches.  

In this paper, we take such issues into consideration by proposing a two-stage multilevel  2D barcode image texture descriptor applied to a novel multilevel 2D barcode. Our approach takes into account a diverse, complementary and robust feature set composed of global and local features, with these features being designed to capture specific distortions suffered by the considered barcode in its counterfeited version. The global features are used to describe artifacts in image pixels, being calculated by the Multiscale Rotation Invariant Binary Gabor Patterns (\texttt{MRIBGP}), whereas the local features are calculated by a Bag of Multiscale Residual Words (\texttt{BOMRW}). A final descriptor fuses both features for a powerful authentication. We validate the effectiveness of our method on two datasets built with a large set of devices. Compared to other descriptors, the results of our extensive experiments showed promising results in challenging scenarios.

The remaining of this paper is organized as follows. Section \ref{proposed_method} discusses the 2D barcode considered and our proposed authentication scheme for 2D Multilevel barcode copy detection. Section \ref{experiments} shows experimental results on diverse scenarios. Finally, Section \ref{conclusion} concludes the work and discusses future perspectives of research.

\begin{figure*}[th!]
	\centering{
		\begin{minipage}[c]{.083\linewidth}
			\centering
			\centerline{\tiny{Genuine}}
			\vspace{2cm}
			\centerline{\tiny{counterfeited}}
		\end{minipage}
		\begin{minipage}[c]{.22\linewidth}
			\centering
			\subfigure{\includegraphics[width=1in,height=1in]{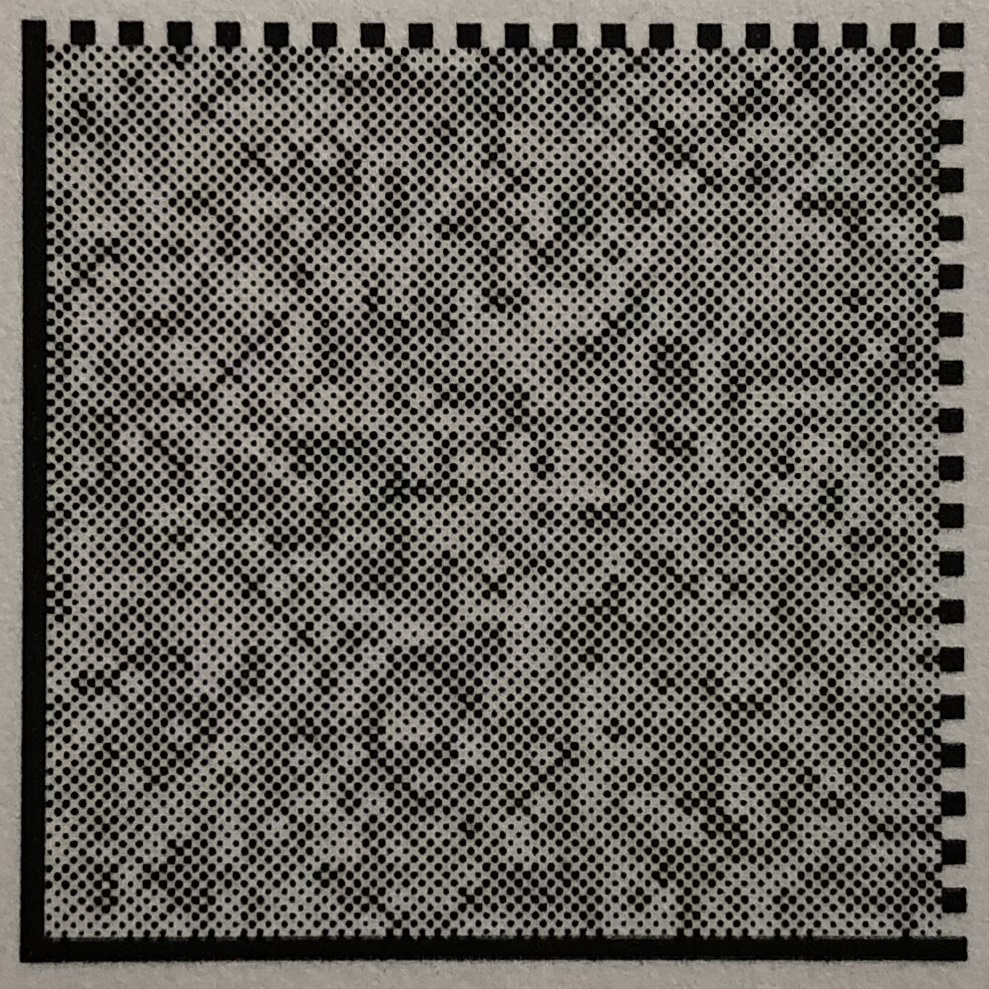}}
			\vspace{0.2cm}
			\subfigure{\includegraphics[width=1in,height=1in]{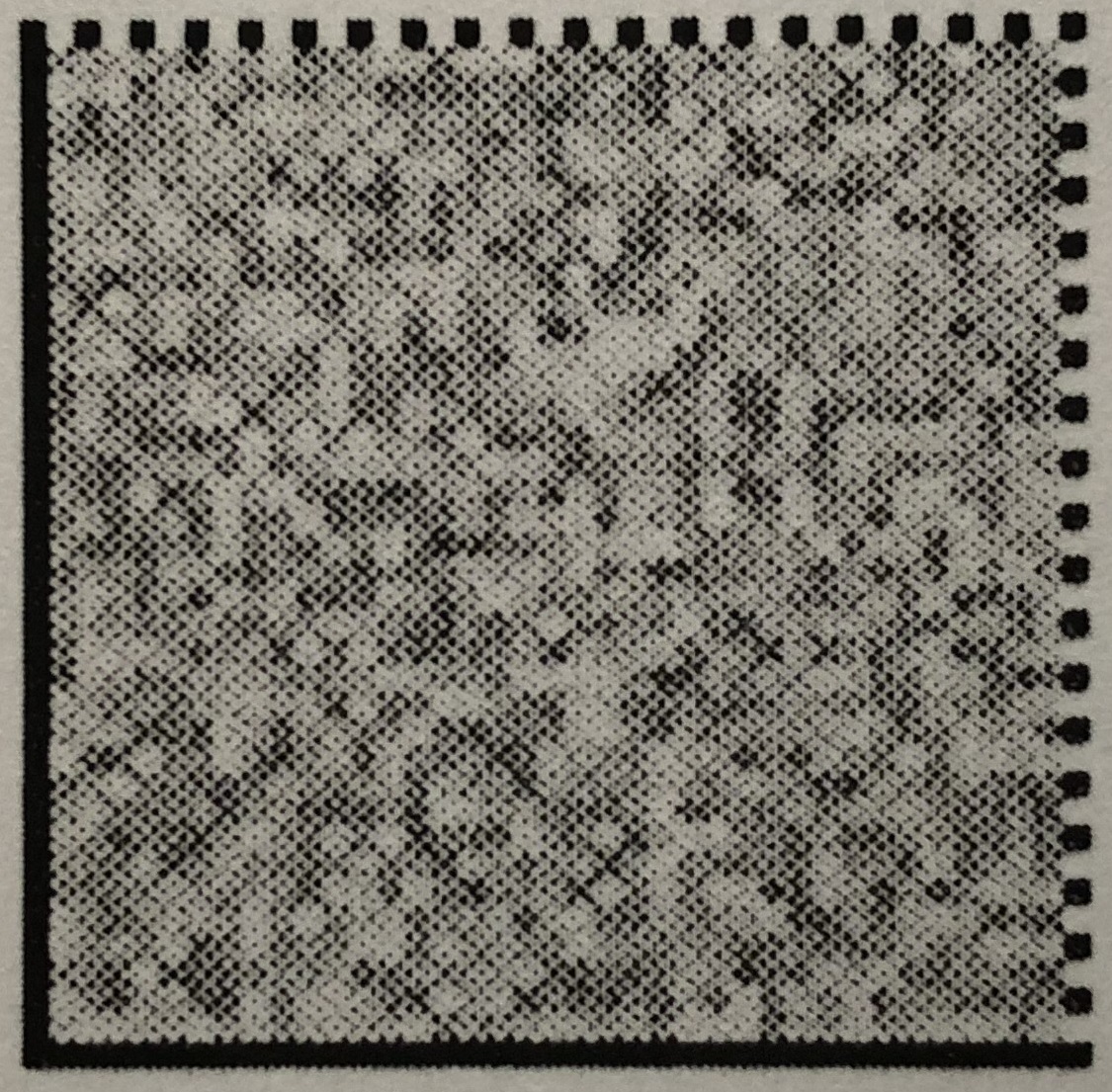}}
			\subfigure{\footnotesize(a) 2.5 $\times$ 2.5 cm$^2$}
		\end{minipage}
		\begin{minipage}[c]{.22\linewidth}
			\centering
			\subfigure{\includegraphics[width=1in,height=1in]{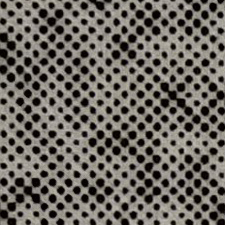}}
			\vspace{0.2cm}
			\subfigure{\includegraphics[width=1in,height=1in]{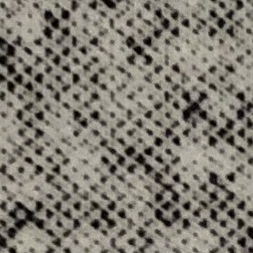}}
			\subfigure{\footnotesize(b) 8$\times$ 8 modules \label{cc}}
		\end{minipage}
		\begin{minipage}[c]{.22\linewidth}
			\centering
			\subfigure{\includegraphics[width=1in,height=1in]{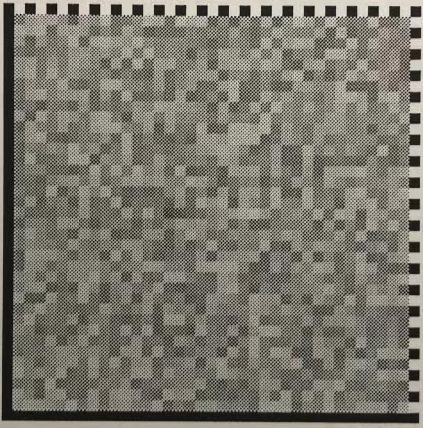}}
			\vspace{0.1cm}
			\subfigure{\includegraphics[width=1in,height=1in]{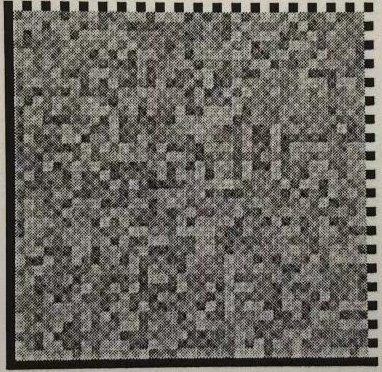}}
			\subfigure{\footnotesize(c) 5 $\times$ 5 cm$^2$ area}
		\end{minipage}
		\begin{minipage}[c]{.22\linewidth}
			\centering
			\subfigure{\includegraphics[width=1in,height=1in]{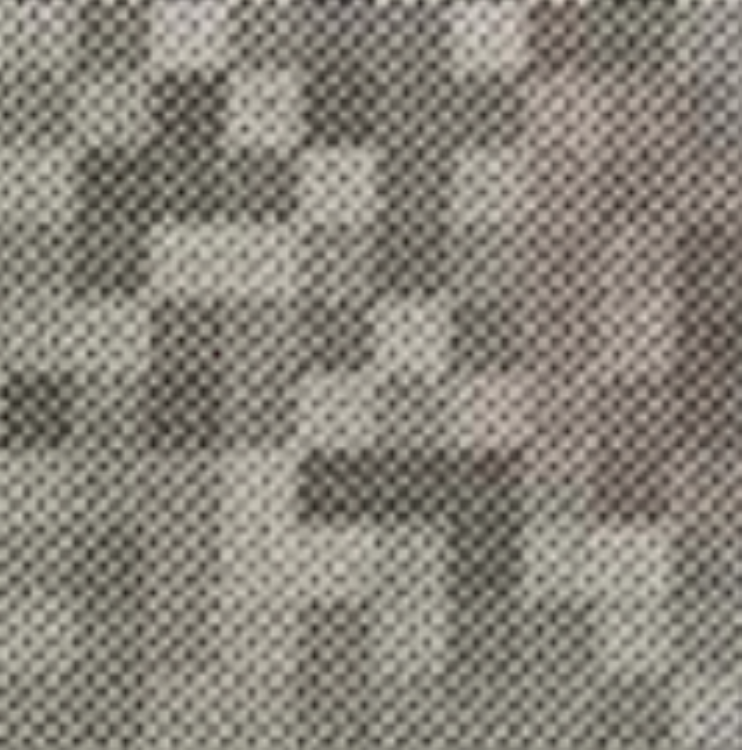}}
			\vspace{0.1cm}
			\subfigure{\includegraphics[width=1in,height=1in]{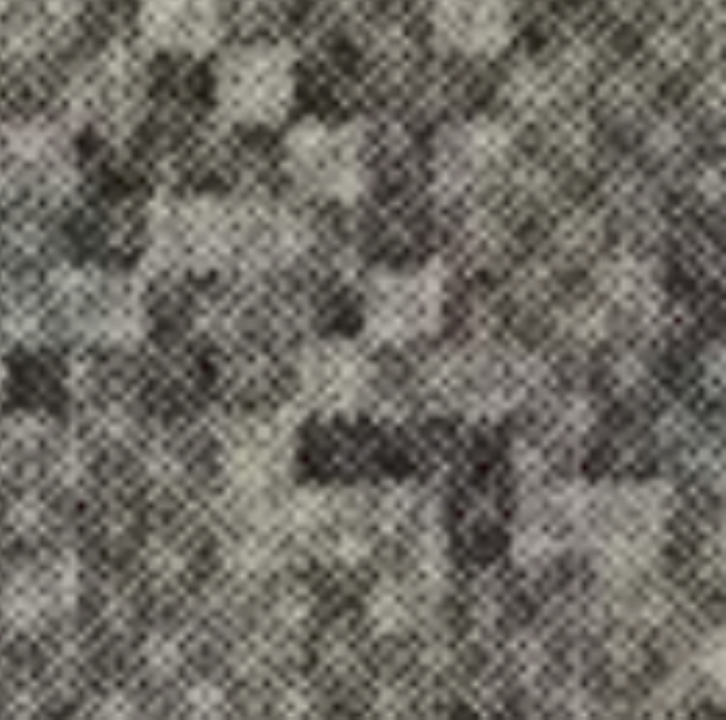}}
			\subfigure{\footnotesize(d) 8$\times$ 8 modules \label{dd}}
		\end{minipage}
	}
	\caption{Illustrations of the spatial artifacts in the genuine and illegal copies of multilevel 2D barcodes after scan and print attack. The top and bottom rows show examples of genuine and counterfeited barcode images respectively, considering different areas and $8\times 8$ , where counterfeited artifacts are distinguishable.}
	\label{fig:BarcodeExampleFeatures}
\end{figure*}

\begin{figure}[t!]
	\centering
	\subfigure[Genuine Patch \label{a}]{\includegraphics[scale=0.27]{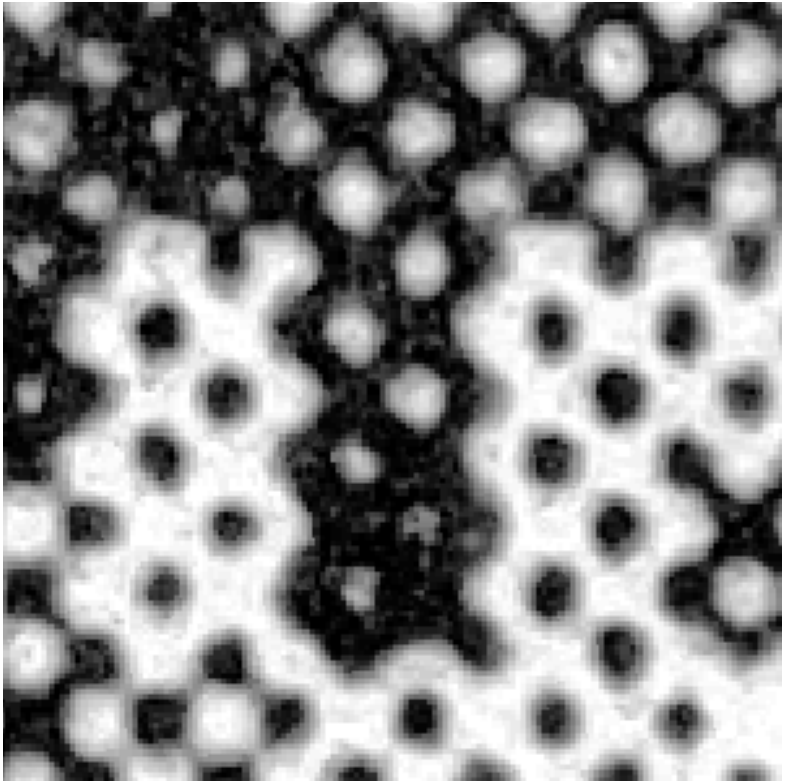}} \hspace{0.2cm}
	\subfigure[Counterfeited Patch\label{b}]{\includegraphics[scale=0.27]{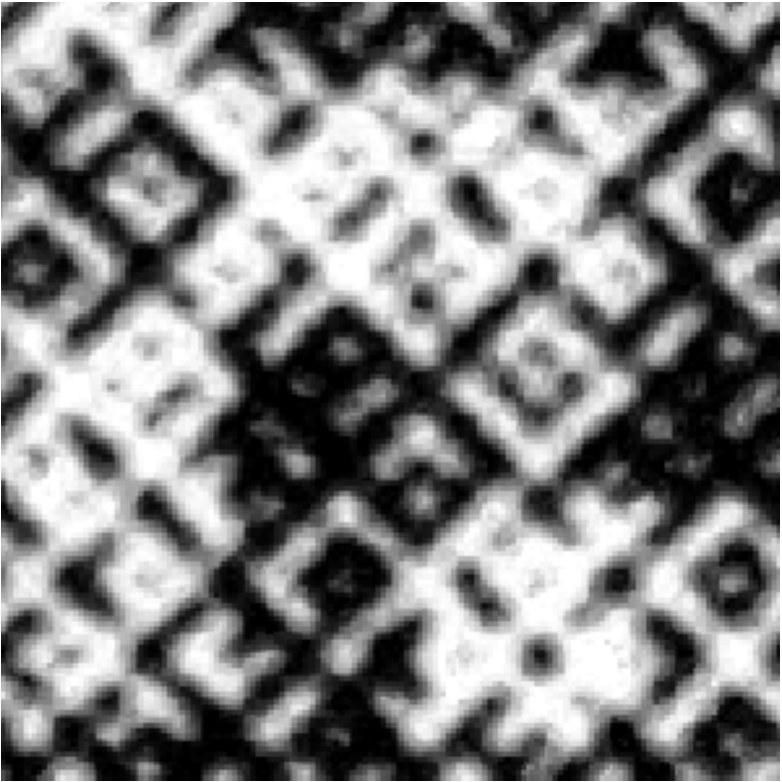}}
	\vfill
	\subfigure[Genuine Patch\label{c}]{\includegraphics[scale=0.93]{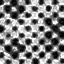}} \hspace{0.2cm}
	\subfigure[Counterfeited Patch\label{d}]{\includegraphics[scale=0.93]{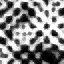}}
	
	\caption{Halftone patterns in small image patches from 2D multilevel barcodes of same size, considering genuine (a-c) and counterfeited (b-d) barcodes. It is clear that the samples in (a) and (b) can be visually discriminated, as the replication artifacts are evident in (b). However, based on the knowledge learned from (a) and (b), it is difficult to differentiate the texture patterns in (c) and (d). Additionally, halftones in the genuine patches (a) and (c) do not have the same appearance. These issues highlight the difficulty of this problem.}
	\label{fig:patches}
\end{figure}

\section{Proposed Method}
\label{proposed_method}

Our solution is demonstrated with the proprietary generic multilevel 2D barcode proposed in \cite{zhang2019Accurate}. Examples of genuine and counterfeited multilevel 2D barcodes are shown in Figure~\ref{fig:BarcodeExampleFeatures}. It can be seen from that figure that the additional scan-and-print operation in the illegal copying process yields specific textural distortions, which differentiate the genuine and counterfeited samples. However, as barcodes can be printed with different sizes and are acquired by different devices on varying angles and distances, the texture patterns can vary significantly even in the same class (\textit{i.e.}, genuine or counterfeited). For example, Figures~\ref{fig:BarcodeExampleFeatures} (a) and (c) illustrate different textures within the same genuine or counterfeited barcodes, but rendered with different printing sizes. Furthermore, Figure~\ref{fig:patches} shows some patches from different genuine and counterfeited barcodes with the same area. It can be seen from Figure~\ref{fig:patches} (b) and (d) that the counterfeited barcodes do not have a common halftone texture pattern. Worse still, the genuine patch in Figure~\ref{fig:patches} (c) has some irregular halftones caused by acquisition noise and the printing process. Therefore, the halftone dots in Figure~\ref{fig:patches} (a) are different from those in Figure~\ref{fig:patches} (c). Such irregular halftoning patterns are not easy to capture with the existing textural descriptors. In this paper, we propose a diverse feature set in order to alleviate such a problem. The proposed features are summarized in the following subsections.

\subsection{Global Feature Set: Multiscale Rotation Invariant Binary Gabor Patterns (MRIBGP)}

The first proposed descriptor is based on \cite{6466800}. The Binary Gabor Pattern (\texttt{BGP}) is calculated by convolving $J$ Gabor filters of different orientations with image patches. The $J$ outputs are the sums of filtering responses in a neighborhood and are concatenated in a response vector $r=\{r{_j}:j=0...J-1\}$. By thresholding the values in $r$, a binary vector $b=\{b{_j}:j=0...J-1\}$ is obtained.
The BGP value is then calculated by converting the $J$-bits binary number in $b$ into decimal. 

To achieve rotation invariance, the maximum \texttt{BGP} is calculated after several circular bitwise right shifts in the binary vector to yield the Rotation Invariant \texttt{BGP} (\texttt{RIBGP}). That is
\begin{equation}
RIBGP=\max\{BGP(\mbox{ROR}(b,j))|j=0,...J-1\},
\end{equation}
where $\mbox{ROR}(b,j)$ is the circular bitwise right shift $j$ times on the $J$-bit number $b$.
\begin{figure*}
\centering
\includegraphics[scale=0.42]{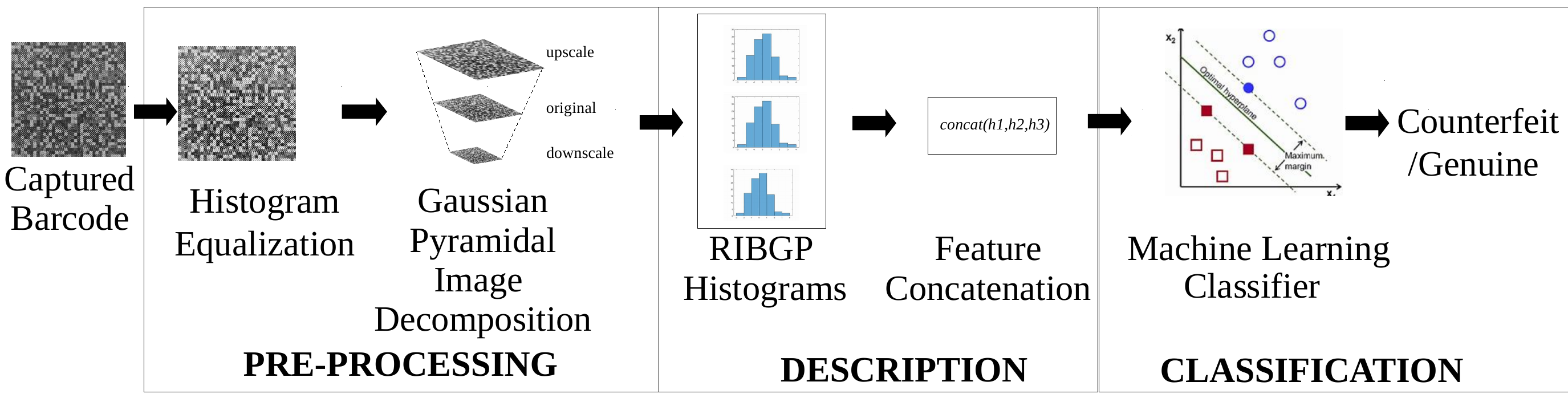}
\caption{The proposed Multiscale Rotation Invariant Binary Gabor Patterns for 2D multilevel barcode authentication.}
\label{fig:pipeline-barcode-texture-descriptor}
\end{figure*}

\begin{figure*}[h!]
	\centering
	\includegraphics[scale=0.37]{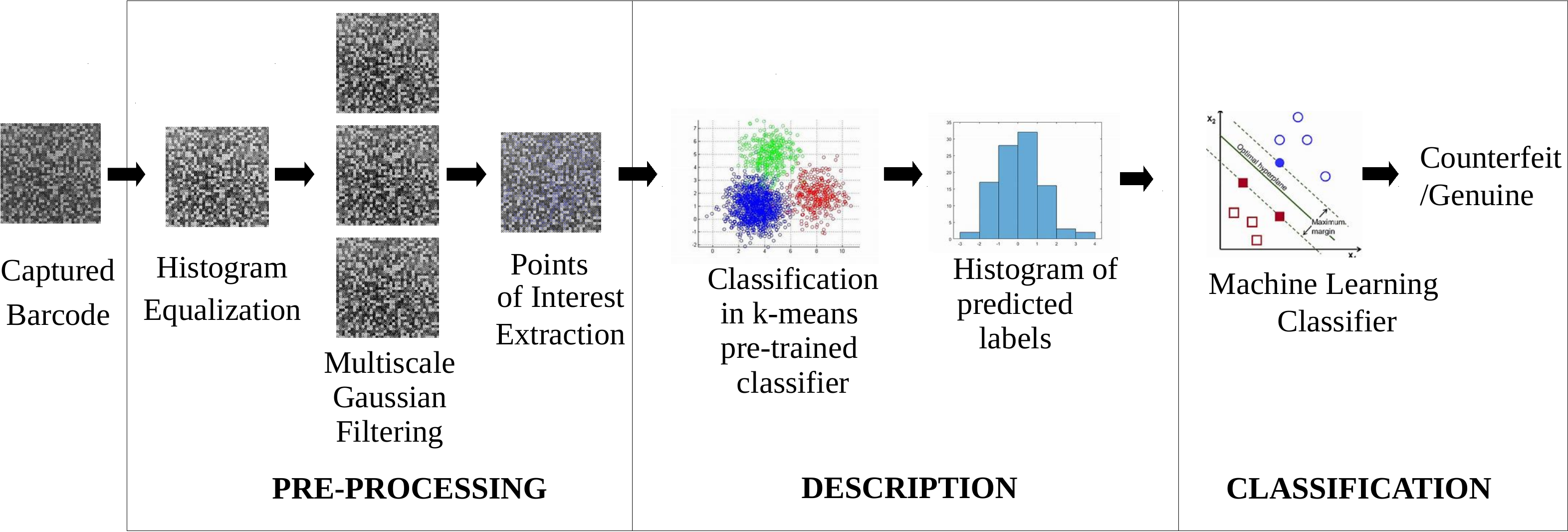}
	\caption{The pipeline of the proposed Bag of Multiscale Residual Words approach for 2D multilevel barcodes authentication}
	\label{fig:pipeline-barcode-surf}
\end{figure*}

By considering $j=8$ (or 8-bit values), there are only 36 unique combinations of maximum \texttt{RIBGP} values to be calculated, which will be the number of bins of a \texttt{RIBGP} values histogram that is built to describe images textures. This approach considers $\gamma=1.82$, three combinations of $\sigma$ and $\lambda$, and eight combinations of $\theta$ to perform texture classification. These parameters generate eight even-symmetric and eight odd-symmetric Gabor filters. Finally, a set of histograms are built for each resolution. Three combinations of $\sigma$ and $\lambda$ are considered, yielding three resolutions per filter symmetry. So, this leads to a $(3+3)\times 36=216$ dimensional feature vector.

In the proposed approach, we aim at analyzing the effect of the spatial frequency artifacts at multiple scales. Our descriptor, called the Multiscale Rotation Invariant Binary Gabor Patterns \texttt{MRIBGP}, applies the histogram of \texttt{RIBGP} values at different levels of the Gaussian Pyramidal decomposed image. In our approach, illustrated in Figure \ref{fig:pipeline-barcode-texture-descriptor}, the 2D barcode image firstly undergoes histogram equalization in order to mitigate the effects of different illumination conditions. Then, a Gaussian Pyramidal Decomposition is performed by downscaling and upscaling the barcode image and the 216-dimensional features are calculated for each scale. After a three-scale decomposition, a 648-dimensional feature vector is extracted and will describe both original and counterfeited barcodes.

\subsection{Local Feature Set: the Bag of Multiscale Residual Words (BOMRW)}
\label{subsec:LocalFeature}

First devised for document classification applications, the Bag of Visual Words (\texttt{BOVW}) is a computer vision technique that describes image contents by occurrence counts in a dictionary of local image features. The \texttt{BOVW} approach was firstly presented in \cite{Csurka04visualcategorization} and it works by detecting points of interest such as \texttt{SIFT} \cite{sift} in training images, which are then clustered by unsupervised classification algorithms to build a dictionary. The size of the visual vocabulary $VOC = \{v_1, v_2, v_3,..., v_k\}$ is $k$, where $k$ is the number of centroids in the clustering process.
A given image can then be represented by a set of descriptors defined as
\begin{equation}
I = \{d_1, d_2, d_3,...,d_N\},
\end{equation}
where $N$ is the total number of descriptors detected for a given image $I$. Each descriptor $d_i$, detected as an $N$-dimensional vector keypoint, is then mapped to a visual word $v_i$.
This is done by finding the minimum Euclidean distance between $d_i$ and each $v_i$. That is
\begin{equation}
v(d_i)=\arg\min_{v \in \mbox{voc}} \mbox{Dist}(v,d_i),
\end{equation}
where Dist$(\cdot)$ evaluates the Euclidean distances between the input vectors.

For our problem, a 2D barcode can be acquired under various conditions with different rendering sizes, acquisition noises, \textit{etc}.
To mitigate these uncontrolled factors, we propose to characterize the points of interest distribution after diverse low-pass filterings of the input image, as described in Figure \ref{fig:pipeline-barcode-surf}. The processing pipeline of our proposed Bag of Multiscale Residual Words (\texttt{BOMRW}) works by first applying histogram equalization to the input 2D barcode image. Then, successive Gaussian filterings of varying filter sizes ($3 \times 3$,  $5 \times 5$ and $7 \times 7$ pixels) are applied to remove image noise and yield a \emph{residual} multi-channel image. Afterwards, keypoints are extracted from the residual image and are clustered through a \textit{k-}means procedure, where $k$ is set as the square root of the number of training samples. Finally, the distances between the cluster centroids and the detected keypoints are calculated to find correspondences. This procedure results in feature vectors (histograms) of visual words for the training images, which can be used to train any machine learning classifier. In the testing stage, the process is repeated, with keypoints being detected and their corresponding visual words found in the pre-trained $k$-means clusters. Then, a histogram of visual words is calculated and used as input to the machine learning classifier, which will be used for the \texttt{BOMRW} based barcode authentication. In our proposed approach, we consider the using the \texttt{BRISK} descriptor \cite{Leutenegger11brisk:binary} to extract keypoints, as it showed better robustness for this specific task than SIFT and SURF.

\subsection{The Merged Feature Set: the Local and Global Multiscale Feature Set (LGMFS)}
\label{subsec:LGMFS}

After both features are calculated, a final feature vector $f^i$  is created by concatenating feature histograms from \texttt{MRIBGP} and \texttt{BOMRW} descriptors in a $(216 \times S) + \sqrt{|T_k|}$ dimensional vector, where $S$ is the number of scales used in the \texttt{MRIBPG} descriptor and $T_k$ is the number of keypoints detected in the training images by \texttt{BOMRW}.
As a next step, we apply L1 normalization to these histograms. We chose L1 normalization as it minimizes the sum of the absolute differences between the target value and its mean. This procedure is of fundamental importance, as it reduces the differences of the range of features resulting from different descriptors. Specifically, the normalized vectors $z_i$ can be written as
\begin{equation}
z^i=\frac{f^i}{\sum_{j=1}^{D} |f^i_j|},
\end{equation}
where $D=(216 \times S) + \sqrt{|T_k|}$ denotes the feature dimension. 

This way, the proposed Local and Global Multiscale Feature Set (\texttt{LGMFS}) merges both the local and global features into one histogram. The resulting histogram will then contain richer information about the genuine and counterfeited barcodes, with descriptive information from edges and keypoints. Such features are then fed to a Support Vector Machines classifier with linear kernel. The LIBSVM library with five-fold cross-validation is adopted in our implementation \cite{chang2011libsvm} to find the best parameters for training the classifier. For reproducibility reasons, the source code of our proposed approach is available at GitHub\footnote{https://github.com/anselmoferreira/2d-barcode-authentication}.

\section{Experiments}
\label{experiments}

In the first experiment, we compare the individual proposed methods against their original counterparts in order to understand how they evolve the original algorithms. For this experiment, we consider the Dataset I described in Table \ref{dataset-1} created to generate genuine and counterfeited barcodes. In this dataset, 30 barcode images are taken under each devices combination, and this procedure leads to 300 genuine barcode images. To create the counterfeited samples, the printed genuine barcodes are first scanned at 600, 1200 and 2400 Points Per Inch (PPI) resolutions, respectively, and then are printed with the same paper of the genuine barcodes. Given a large amount of counterfeiting devices combinations, only some representative ones have been used to generate the counterfeited barcodes. This way, Dataset I includes 3775 counterfeited barcode images, totaling $3775+300=4075$ images.

\begin{table}[htbp]
\caption{Dataset I devices used to generate  $2.5 \times 2.5 \mbox{ cm}^2$ area barcodes.}
\centering
\scriptsize
\begin{tabular}{|c|c|c|c|}
\hline
\textbf{Type} & \textbf{Brand} & \textbf{Model}             & \textbf{Resolution} \\
Smartphone & Google          & Nexus4       & 8MP \\
Smartphone & Apple         & Iphone 8 Plus       & 12MP \\
Smartphone & Huawei      & P9       & 12MP \\
Smartphone & Meizu          & Metal       & 13MP \\
Smartphone & Nokia          & Lumia 930       & 20MP \\
Printer & HP          & Laserjet Pro M401dn       & 600 and 1200dpi \\
Printer & Toshiba     & E-studio 2307       & 600dpi \\
Scanner & Epson     & Perfection V330       & 600, 1200 and 2400dpi \\
Scanner & Microtek     & D328K       & 600, 1200 and 2400dpi \\
Scanner & BenQ     & K810      & 600, 1200 and 2400dpi \\
\hline
\end{tabular}
\label{dataset-1}
\end{table}

\begin{figure}[h!]
	\centering
	\includegraphics[width=.6\linewidth]{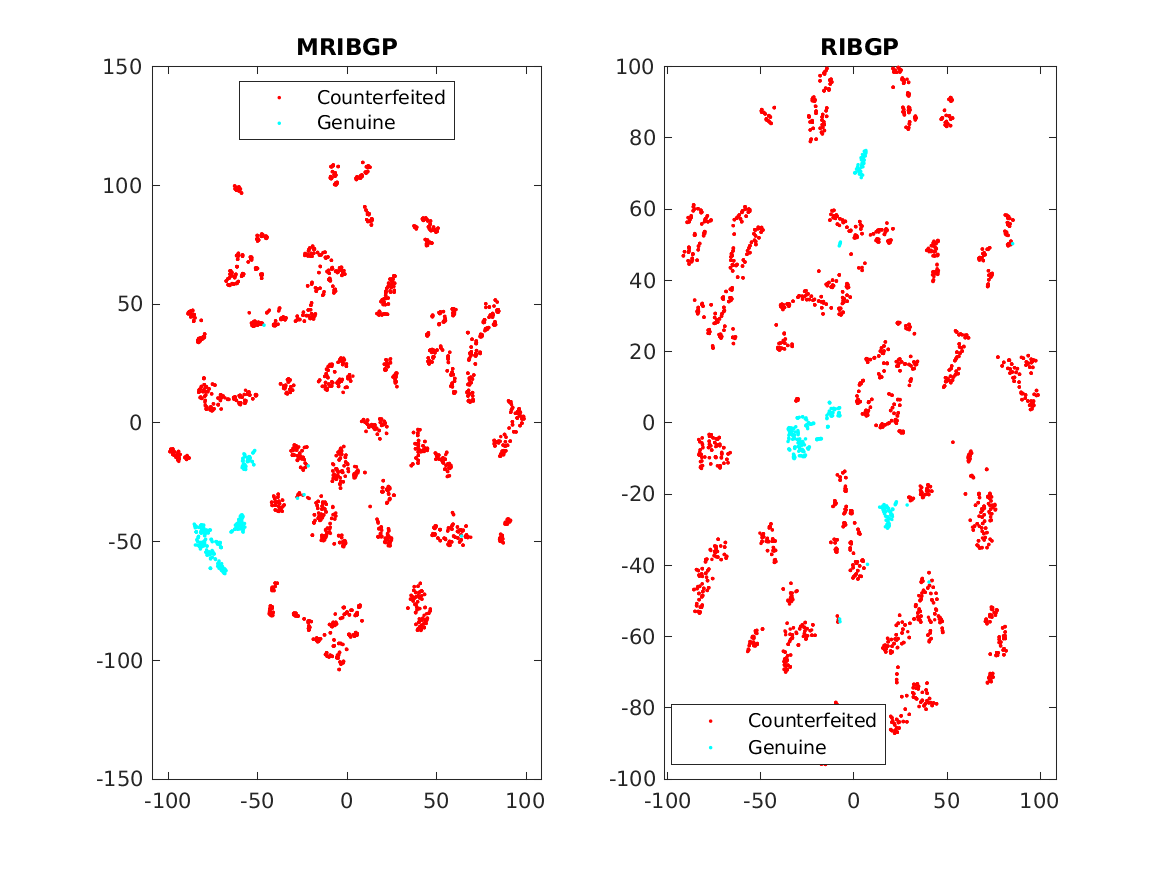}
	\caption{Clustered features from the proposed Multiscale Rotation Invariant Binary Gabor Patterns (left) versus those from the original Rotation Invariant Binary Gabor Patterns \cite{6466800} (right). It can be seen that the proposed features cause less confusion, as the clusters from genuine barcodes are closer to each other and are farther from the counterfeited barcode clusters. This behavior shows that samples described by the proposed features are better separable by a classifier.}
	\label{fig:features-mribgp}
\end{figure}

Figure~\ref{fig:features-mribgp} plots the clustered features of the proposed MRIBGP versus the original RIBGP, using the t-SNE visualization tool \cite{maaten2008visualizing} after both descriptors are applied to dataset I. On the one hand, it can be seen that the RIBGP descriptor with only one scale leads to more confusion in the feature clustering. This is because the artifacts and halftones with different sizes have not been considered in the feature description. On the other hand, the advantage of the proposed multiscale approach is achieved by employing pyramidal decomposition in the input images, which transforms the halftones into multiple sizes and minimizes this way noisy effects present in both  genuine and counterfeited captured samples.

\begin{figure}[h!]
	\centering
	\includegraphics[width=.6\linewidth]{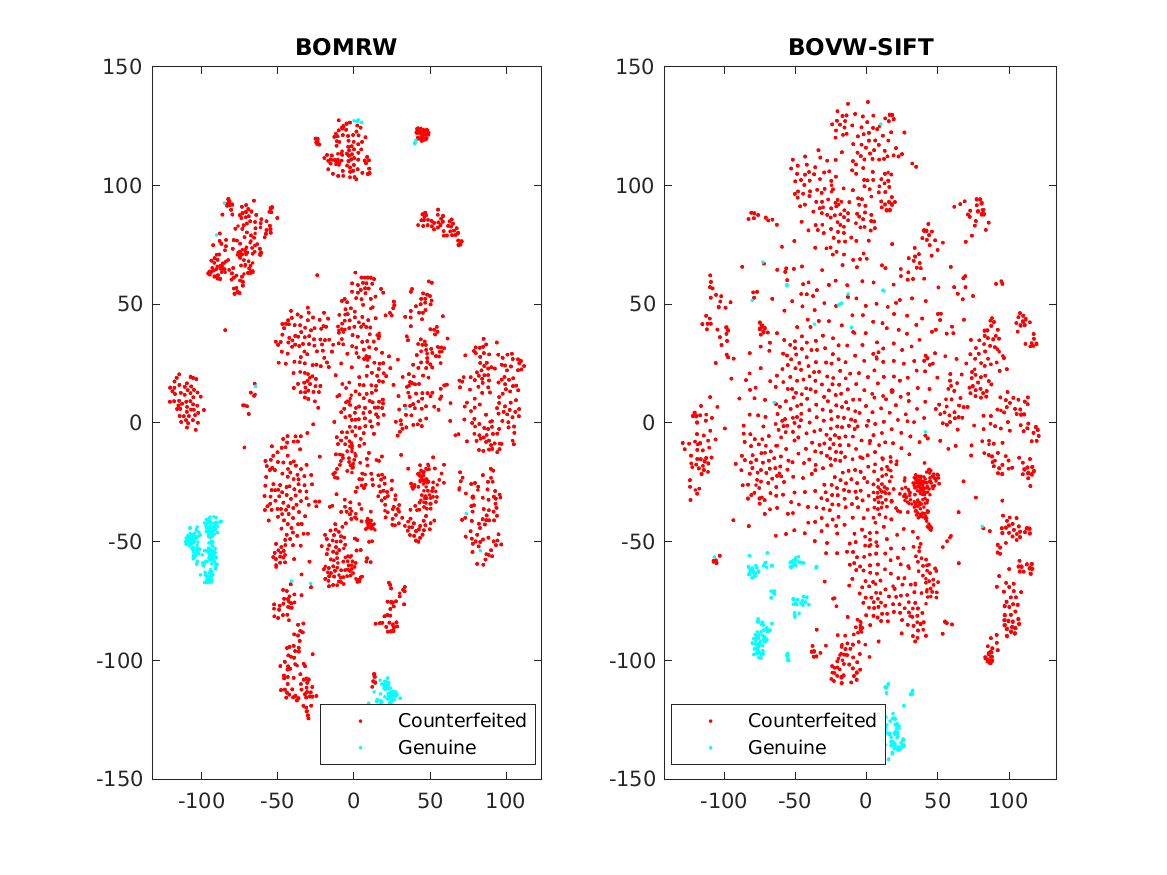}
	\caption{Clustered features from the proposed Bag of Multiscale Residual Words with the BRISK descriptor (left) versus those from the original Bag of Visual Words using SIFT descriptor (right). It can be seen that the proposed features cause less confusion as the clusters are less spread out in the feature space and are farther from the counterfeited barcode features clusters.}
	\label{fig:pipeline-clustermborw}
\end{figure}

Figure \ref{fig:pipeline-clustermborw} now shows the t-SNE visualizations of feature vectors from the proposed BOMRW and the traditional BOVW approach using SIFT descriptor when both descriptors are applied to some 2D barcodes. It can be noticed from this figure that the proposed feature set tends to cluster genuine samples in two main clusters, with few samples close to the counterfeited clusters. For the traditional BOVW, the genuine samples are spread out in multiple clusters which are far away from each other, but are close to the counterfeited class clusters. This issue seriously hinders the authentication performance, as these clusters setup may confuse classifiers. Such a behavior will be seen affecting experiments results of such a descriptor in the remaining of this paper.

We now validate our approaches in a realistic setup, where two sets of different devices are allowed to generate the training and testing samples. For that, we use the Dataset I already presented in Table \ref{dataset-1} for training, and a new (Dataset II) presented in Table \ref{dataset-2} to generate new barcodes of same area. This new configuration leads to 120 genuine and 960 counterfeited barcode pictures, totalling $960+120=1080$ images for Dataset II. In this scenario we also invert the training/testing order, using Dataset II for training and Dataset I for testing. We use as metrics the f-measure (F), Normalized Accuracy (NACC), True Positive Rate (TPR) and False Positive Rate (FPR). Mean results of these two experiments are reported in Table \ref{results-crossdataset}.

\begin{table}[htbp]
\caption{Dataset II devices used to generate  $2.5 \times 2.5 \mbox{ cm}^2$ area barcodes.}
\centering
\scriptsize
\begin{tabular}{|c|c|c|c|}
\hline
\textbf{Type} & \textbf{Brand} & \textbf{Model}             & \textbf{Resolution} \\
Smartphone & Apple  & Iphone 6       & 8MP \\
Smartphone & Xiaomi & Mi-25       & 8MP \\
Printer & Ricoh & Aficio MP C5000       & 1200dpi \\
Printer & Ricoh & Aficio MP C3501       & 1200dpi \\
Scanner &Canon & CanoScan Lide       & 600 and 1200dpi \\
Scanner & HP & LaserJet-Pro   & 600 and 1200dpi \\
\hline
\end{tabular}
\label{dataset-2}
\end{table}

\begin{table}[htbp]
\caption{Cross dataset experiment results.  Proposed approaches are boldfaced, results are ordered by f-measure and best results per metric are highlighted in gray.}
\centering
\scriptsize
\begin{tabular}{|c|c|c|c|c|}
\hline
\textbf{Approach} & \textbf{F} & \textbf{NACC}  & \textbf{TPR} & \textbf{FPR} \\
\textbf{\texttt{LGMFS}}&	\cellcolor{lightgray}0.97$\pm$0.04&	\cellcolor{lightgray}96.48$\pm$3.16&	\cellcolor{lightgray}98.58$\pm$0.82&	5.62$\pm$7.15 \\
\textbf{\texttt{BOMRW}}&	0.96$\pm$0.04&	92.57$\pm$1.30&	93.31$\pm$9.46&	8.16$\pm$6.83 \\
\texttt{BOVW-SIFT}\cite{Csurka04visualcategorization} &	0.96$\pm$0.03 &	91.04$\pm$1.35 & 94.66$\pm$7.55&	12.58$\pm$4.82 \\
\texttt{RESNET}\cite{he2016deep} &	0.95$\pm$0.04 &	93.42$\pm$0.83 & 90.84$\pm$7.33&	4.00$\pm$5.65 \\
\texttt{DENSENET}\cite{huang2017densely}  &	0.95$\pm$0.02 &	93.40$\pm$0.35 & 90.13$\pm$5.41&	3.33$\pm$4.71 \\
\textbf{\texttt{MRIBGP}} & 0.86$\pm$0.13 & 97.64$\pm$2.00 & 95.94$\pm$4.94 & \cellcolor{lightgray}0.66$\pm$0.94 \\
\texttt{RIBGP}\cite{zhang2019Accurate} &	0.83$\pm$0.19&	96.49$\pm$3.25&	94.40$\pm$7.34&	1.41$\pm$0.82 \\
\hline
\end{tabular}
\label{results-crossdataset}
\end{table}

From results shown in Table~\ref{results-crossdataset}, we find it worth starting discussing the CNNs performance in this more difficult scenario. These approaches showed better results only when trained from scratch and yield only reasonable authentication results, as their performance is highly dependent on the uniformity of training and testing sets and also the number of training data, which can be limited due to high costs of building such datasets. Such limitation results in an unacceptable false negative rate (\emph{i.e.} percentage of undetected counterfeited samples) of almost 10\% in all CNNs evaluated in this experiment.

Results in Table~\ref{results-crossdataset} also highlight the improvement of the proposed \texttt{MRIBGP} over the original \texttt{RIBGP} descriptor \cite{6466800}. The multi-scale transform and illumination correction from the proposed approach generate some clean samples to the descriptor, which leads to a better description of the counterfeited samples. Therefore, the proposed \texttt{MRIBGP} descriptor improves all the metrics of the original \texttt{RIBGP} approach \cite{6466800}. Specifically, the genuine barcodes detection performance has been improved considerably, which is highlighted by the lowest mean false counterfeited detection (FPR) of this experiment. This supports our hypothesis that the rotation and multi-scale invariant descriptors can be crucial for the problem of multilevel 2D barcodes authentication, as the patterns are irregular due to the use of different devices and acquisition conditions. Similarly, the proposed bag of visual words approach (\texttt{BOMRW}) has the second best result of this experiment, being thus a better option for this problem than using the common approach with the SIFT descriptor (\texttt{BOVW-SIFT}), as the proposed residual image highlights better the distortions of irregular edges in counterfeited images. Finally, the effect of combining both descriptors in a final descriptor based on edges and points of interest in the proposed \texttt{LGMFS} leads to a better performance, especially because both  fused descriptors take into the account varying halftones and edge behaviors, but considering different image structural features (\textit{e.g.}, pixels and keypoints).
The \texttt{LGMFS} descriptor shows the best f-score of 0.97, 96.48\% NACC and the best TPR of 98.58\%. Therefore, our approach, even taking into account less data, showed a significantly better result than data-hungry approaches like CNNs.


We finish the validation experiments with an even more difficult scenario. In this experiment, the robustness of the proposed scheme under different printing area barcodes is investigated. To do that, we test the approaches on a $5 \times 5$ cm$^2$ area barcodes Dataset III described in Table \ref{dataset-3}. This new dataset contains 180 images, including 60 genuine and 120 counterfeited barcodes.  Table \ref{results-crosssize} shows the performance of descriptors considered in this experiment.

\begin{table}[htbp]
\caption{Dataset III devices used to generate $5 \times 5$ cm$^2$ area barcodes.}
\centering
\scriptsize
\begin{tabular}{|c|c|c|c|}
\hline
\textbf{Type} & \textbf{Brand} & \textbf{Model}             & \textbf{Resolution} \\
Smartphone & Apple         & Iphone 8 Plus       & 12MP \\
Printer & HP          & Laserjet Pro M401dn       & 600 and 1200dpi \\
Scanner & Epson     & Perfection V330       & 600, 1200 and 2400dpi \\
\hline
\end{tabular}
\label{dataset-3}
\end{table}

\begin{table}[htbp]
\caption{Cross area experiment results. Proposed approaches are boldfaced, results are ordered by f-measure and best results per metric are highlighted in gray.}
\centering
\scriptsize
\begin{tabular}{|c|c|c|c|c|}
\hline
\textbf{Approach} & \textbf{F} & \textbf{NACC}  & \textbf{TPR} & \textbf{FPR} \\
\textbf{\texttt{LGMFS}} & \cellcolor{lightgray}0.99 & \cellcolor{lightgray}98.33 & \cellcolor{lightgray}100.00 & 3.33\\
\textbf{\texttt{MRIBGP}} & 0.98& 97.08 & 99.16  & 5.00 \\
\texttt{DENSENET} \cite{huang2017densely} & 0.96 & 92.50 &  \cellcolor{lightgray}100.00 & 15.00\\
\textbf{\texttt{BOMRW}} & 0.93 & 93.75 & 87.50 & \cellcolor{lightgray} 0.00\\
\texttt{BOVW-SIFT+RIBGP} & 0.93 & 85.83 & \cellcolor{lightgray}100.00   & 28.33  \\ 
\texttt{RIBGP} \cite{zhang2019Accurate} & 0.92 & 83.33 & \cellcolor{lightgray}100.00 & 33.33\\
\texttt{RESNET} \cite{he2016deep} & 0.80 & 52.50 & \cellcolor{lightgray}100.00 & 95.00\\
\texttt{BOVW-SIFT} \cite{Csurka04visualcategorization} & 0.78 & 82.08 & 64.17 & \cellcolor{lightgray} 0.00\\ 
\hline
\end{tabular}
\label{results-crosssize}
\end{table}

\begin{figure*}[t!]
	\centering
	\subfigure{\includegraphics[scale=0.2]{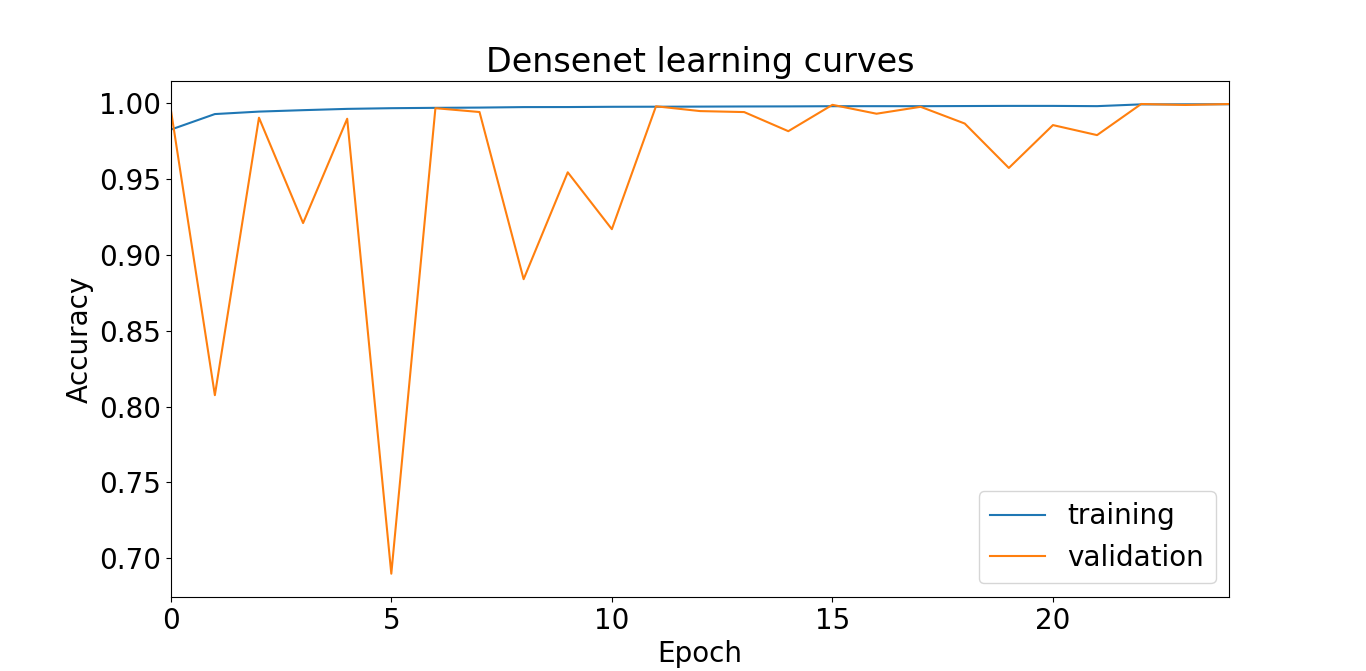}}
	\qquad
	\subfigure{\includegraphics[scale=0.2]{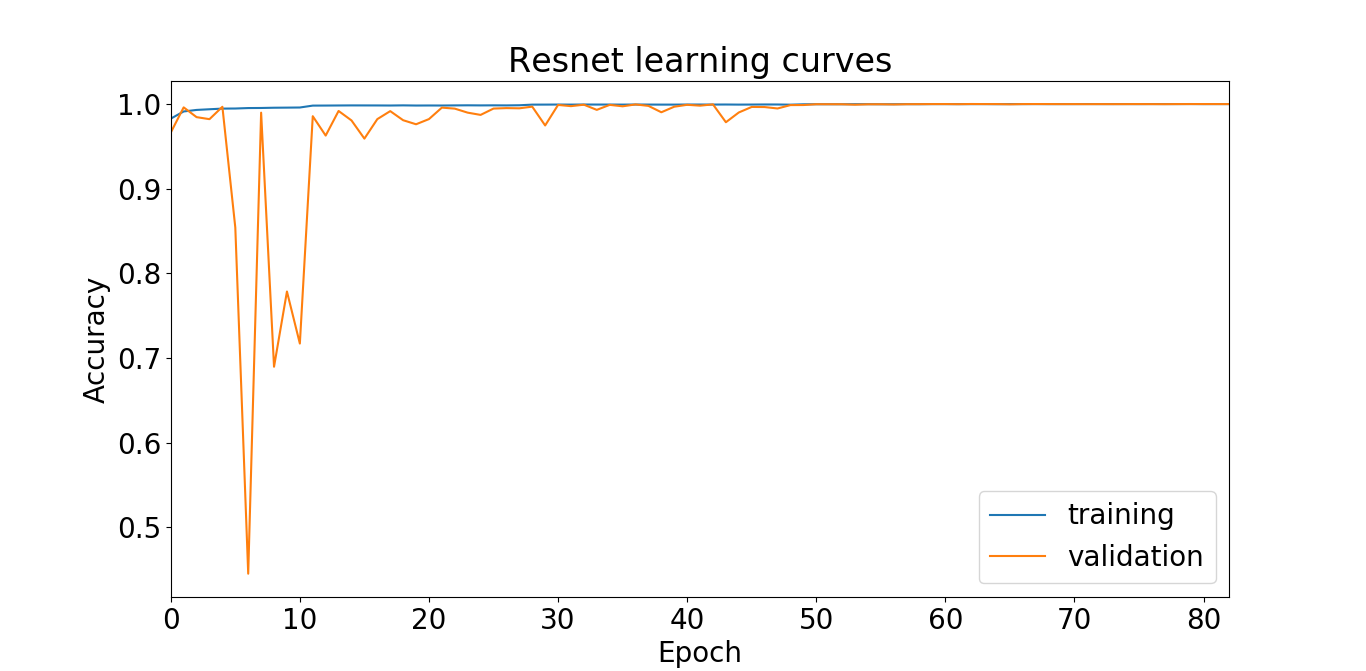}}
	\caption{Training and validation accuracies of the baseline CNNs \cite{huang2017densely, he2016deep}. The high and stable learning curves highlight the fact that the CNNs understand very well the pristine and counterfeited patterns within the same dataset.}
\label{cnns_learning_curves}
\end{figure*}

First, the experimental results show that shallower and simpler CNNs, such as the \texttt{RESNET} with 50 layers, do not perform well in the cross-size evaluation although the training and validation processes have been carried out properly, as can be seen in Figure \ref{cnns_learning_curves}. This can be explained by the fact that the testing dataset has different halftone dots distributions than the ones found in the training set, making it difficult for such CNN to generalize to such a new dataset. We identified that original barcode images with larger printing areas can be easily misclassified as forgeries by such CNN when its trained with patches from smaller size barcodes.
A reason for such a behavior is the fact that more halftone dots are found in the barcode with a larger printing area, and these dots are confused with replication artifacts (\textit{e.g.}, extra edges caused by distortion) in the counterfeited barcodes from Dataset I (with smaller barcode areas), which were used to train the CNNs. However, \texttt{DENSENET} with 121 layers \cite{huang2017densely} can handle that problem as it consists of a large number of layers and dense modules that better generalize to such a new testing dataset. With these features in such a CNN, complex interdependencies between different higher level features can be abstracted.

Finally, Table~\ref{results-crossdataset} also highlights the benefits of joining the local and global descriptors also when considering the original algorithms. Fusing the \texttt{RIBGP} and \texttt{BOVW-SIFT} improved the best individual algorithm performance in 2.50\% when considering the NACC. Notwidthstanding, the scale invariance property of the proposed \texttt{MRIBGP}, which achieves an almost perfect NACC, improves the original \texttt{RIBGP} descriptor by a large margin. The proposed \texttt{BOMRW} descriptor also improves its original counterpart \texttt{BOVW-SIFT} \cite{Csurka04visualcategorization} in such a challenging scenario. Finally, the combination of the proposed descriptors (\texttt{LGMFS}) achieves 98.33\% NACC and 0.99 f-measure. Fusing such descriptors is particularly important when one set of features complements the other to describe unstable patterns, which happens in both experiments reported in this paper.

\section{Conclusion}
\label{conclusion}

In this paper, we proposed a copy-proof 2D barcode scheme which is applicable to a generic multilevel 2D barcode. After investigating the variation of textural appearances in both genuine and counterfeited barcodes, we presented an authentication scheme based on global and local features which minimizes such effects. Extensive experiments have been performed with varying levels of difficulty to demonstrate the advantages of the proposed method. Some promising directions for future research are listed in the following. First, the study of CNNs modules and architectures robust to the replication artifacts of 2D barcodes under different acquisition conditions and barcode sizes is a natural extension of this research. Second, the proposed authentication scheme can be extended to the analysis of halftones separately, instead of pixels values and points of interest. Finally, investigating the proposed descriptors performance in anti-copying problems of color/text documents printed with some general halftoning techniques is of paramount interest.



{\footnotesize
\bibliographystyle{ieee}
\bibliography{references}

\begin{thebibliography}{10}\itemsep=-1pt

\bibitem{chang2011libsvm}
C.-C. Chang and C.-J. Lin.
\newblock {LIBSVM: a Library for Support Vector Machines}.
\newblock {\em ACM Transactions on Intelligent Systems and Technology (TIST)},
  2(3):27, 2011.

\bibitem{8794824}
C.~{Chen}, M.~{Li}, A.~{Ferreira}, J.~{Huang}, and R.~{Cai}.
\newblock A copy-proof scheme based on the spectral and spatial barcoding
  channel models.
\newblock {\em IEEE Transactions on Information Forensics and Security},
  15:1056--1071, 2020.

\bibitem{Csurka04visualcategorization}
G.~Csurka, C.~R. Dance, L.~Fan, J.~Willamowski, and C.~Bray.
\newblock Visual categorization with bags of keypoints.
\newblock In {\em European Conference on Computer Vision}, pages 1--22, 2004.

\bibitem{icc2016}
{Frontier Economics}.
\newblock The economic impacts of counterfeiting and piracy.
\newblock Technical report, {Frontier Economics}, 2016.

\bibitem{he2016deep}
K.~He, X.~Zhang, S.~Ren, and J.~Sun.
\newblock {Deep Residual Learning for Image Recognition}.
\newblock In {\em Proceedings of the IEEE Conference on Computer Vision and
  Pattern Recognition}, pages 770--778, 2016.

\bibitem{huang2017densely}
G.~Huang, Z.~Liu, L.~Van Der~Maaten, and K.~Q. Weinberger.
\newblock Densely connected convolutional networks.
\newblock In {\em IEEE Conference on Computer Vision and Pattern Recognition
  (CVPR)}, pages 2261--2269, 2017.

\bibitem{Leutenegger11brisk:binary}
S.~Leutenegger, M.~Chli, and Y.~Siegwart.
\newblock Brisk: Binary robust invariant scalable keypoints.
\newblock In {\em IEEE International Conference on Computer Vision}, pages
  2548--2555, 2011.

\bibitem{sift}
D.~Lowe.
\newblock Distinctive image features from scale-invariant keypoints.
\newblock {\em International Journal of Computer Vision}, vol. 60(2):91--110,
  2004.

\bibitem{maaten2008visualizing}
L.~v.~d. Maaten and G.~Hinton.
\newblock Visualizing data using t-sne.
\newblock {\em Journal of machine learning research}, 9(Nov):2579--2605, 2008.

\bibitem{ONO2016991}
S.~Ono, T.~Maehara, and K.~Minami.
\newblock Coevolutionary design of a watermark embedding scheme and an
  extraction algorithm for detecting replicated two-dimensional barcodes.
\newblock {\em Applied Soft Computing}, 46:991 -- 1007, 2016.

\bibitem{tkachenko2016two}
I.~Tkachenko, W.~Puech, C.~Destruel, O.~Strauss, J.-M. Gaudin, and C.~Guichard.
\newblock {Two-level QR code for Private Message Sharing and Document
  Authentication}.
\newblock {\em IEEE Transactions on Information Forensics and Security},
  11(3):571--583, 2016.

\bibitem{voloshynovskiy2016physical}
S.~Voloshynovskiy, T.~Holotyak, and P.~Bas.
\newblock {Physical Object Authentication: Detection-theoretic Comparison of
  Natural and Artificial Randomness}.
\newblock In {\em IEEE International Conference on Acoustics, Speech and Signal
  Processing (ICASSP)}, pages 2029--2033. IEEE, 2016.

\bibitem{zhang2019Accurate}
L.~Zhang, C.~Chen, and W.~H. Mow.
\newblock {Accurate Modeling and Efficient Estimation of the Print-Capture
  Channel With Application in Barcoding}.
\newblock {\em {IEEE Transactions on Image Processing}}, 28(1):464--478, 2019.

\bibitem{6466800}
L.~Zhang, Z.~Zhou, and H.~Li.
\newblock Binary gabor pattern: An efficient and robust descriptor for texture
  classification.
\newblock In {\em IEEE International Conference on Image Processing}, pages
  81--84, Sept 2012.

\end{thebibliography}
}

\end{document}